\documentclass[runningheads]{llncs}
\usepackage{graphicx}
\usepackage{amsmath}
\usepackage{url}
\usepackage[tableposition=top]{caption}
\usepackage{float}
\floatstyle{plaintop}
\restylefloat{table}

\usepackage{array}
\newcolumntype{x}[1]{>{\centering\let\newline\\\arraybackslash\hspace{0pt}}p{#1}}
\let\ACMmaketitle=\maketitle
\renewcommand{\maketitle}{\begingroup\let\footnote=\thanks \ACMmaketitle\endgroup}
\begin{document}
\title{A Feature Selection Method for Multi-Dimension Time-Series Data\footnote{The final authenticated version is available online at \protect\url{https://doi.org/10.1007/978-3-030-65742-0_15}}
}
\titlerunning{Feature Selection for Time-Series Data}
%

%

\author{Bahavathy Kathirgamanathan\inst{1} \and
P\'adraig Cunningham\inst{1}}

\authorrunning{B. Kathirgamanathan, P. Cunningham}
\institute{School of Computer Science, University College Dublin\\
\email{bahavathy.kathirgamanathan@ucdconnect.ie}\
}
\maketitle              
\begin{abstract}
Time-series data in application areas such as motion capture and activity recognition is often multi-dimension. In these application areas data typically comes from wearable sensors or is extracted from video. There is a lot of redundancy in these data streams and good classification accuracy will often be achievable with a small number of features (dimensions). In this paper we present a method for feature subset selection on multidimensional time-series data based on mutual information. This method calculates a merit score (MSTS) based on correlation patterns of the outputs of classifiers trained on single features and the `best' subset is selected accordingly.
MSTS was found to be significantly more efficient in terms of computational cost while also managing to maintain a good overall accuracy when compared to Wrapper-based feature selection, a feature selection strategy that is popular 
elsewhere in Machine Learning. We describe the motivations behind this feature selection strategy and evaluate its effectiveness on six time series datasets. 

\keywords{Time-Series Classification  \and Feature Selection \and Merit Score}
\end{abstract}
\section{Introduction}
Multi-dimension time-series data
arises in various application areas such as motion capture and activity recognition \cite{Johnston2018,OReilly2018wearable}. This data will often contain a lot of redundancy with some of the data streams being highly correlated. For this reason, it is important to be able to identify a subset of the features (data streams) that is adequate to characterize the phenomenon under investigation. This is a special case of the feature selection problem in Machine Learning (ML) but in this case the `feature' is a complete time-series rather than a feature in a feature vector representation.  
\par
Time-series data is often not compatible with the standard ML feature selection strategies. Filter strategies are not directly applicable due to the nature of the data and Wrapper methods can be computationally prohibitive (see section \ref{sec:FS} for more detail). 
\par
In this paper, a feature subset selection method for multivariate time series is implemented with the aim of identifying the optimal feature subset to use for classification. The method uses feature-feature correlations as well as feature-class correlations based on mutual information (MI) which are then used to calculate a merit score for each feature subset which will act as the basis upon which to select the `best' subset. The main novelty is that these correlations are calculated on 
the outputs of classifiers trained on single features rather than on the time-series data. 
\par
The following section of this paper presents an overview of existing feature selection techniques. Section 3 describes the Merit Score based technique used for time series (MSTS), Section 4 presents our evaluation of MSTS on selected datasets, and finally Section 5 discusses the conclusions and scope for further work.
\section{Feature Selection}\label{sec:FS}

In a data set of $n$ dimensions there are $2^n$ possible feature subsets.
Feature Selection techniques  explore this space of feature subsets to find the `best' subset. Evaluation strategies can be divided into two broad categories:
\begin{itemize}
\item[--] \textbf{Filter} methods use an external measure such as information gain or a $\chi^2$ statistic to score the informativeness of features. Then a selection criterion will determine the best features to select according to this score, e.g. select features scoring above a threshold or select the top $m$ features. 

\item[--] \textbf{Wrapper} methods for feature selection make use of the learning algorithm itself to choose a set of relevant features. The Wrapper conducts a search through the feature space, evaluating candidate feature subsets by estimating the predictive accuracy of the classifier built on that subset. The goal of the search is to find the subset that maximises this criterion.
\end{itemize}
Filter methods are not computationally expensive but are less accurate as features are not evaluated in context. Wrapper methods can be very effective because they evaluate what is important, the classification performance of different feature subsets. However, because of the extent and nature of the evaluation, Wrappers are computationally expensive. 
\par

\subsection{Correlation based feature selection using Mutual Information}\label{sec:MI}
Correlation based feature selection (CFS) is a compromise between Filter and Wrapper methods as it evaluates features in context but using correlation rather than classification accuracy \cite{hall1999correlation}. CFS is the default feature selection method in Weka \cite{frank2009weka} and has been widely used. However, CFS is not usable with time-series data because it requires data in a feature vector format.
CFS assigns a merit score $M_S$ to a feature subset as follows:
\begin{equation}
    M_S = \frac{k\overline{r_{cf}}}
    {\sqrt{k+k(k-1)\overline{r_{ff}}}}
    \label{eqn:MS}
\end{equation}
Where $\overline{r_{cf}}$ is the average correlation between the features in the subset and the class label and $\overline{r_{ff}}$ is the average correlation between the selected features. $k$ represents the number of features in the subset. These correlations have been measured using techniques such as symmetrical uncertainty based on information gain, feature weighting based on the Gini-index, and a method using the minimum description length (MDL) principle \cite{hall1999correlation}. Information gain based methods have worked well previously and hence the correlations in this paper will be measured using Mutual Information (MI). 
MI has been widely used and has produced successful results for feature selection \cite{Doquire2012}. Generally, as the MI between two random variables increases, the greater the correlation between them will be.
\par
MI is a concept that is used widely in information theory and is based on Shannon's entropy \cite{Shannon1948}, which is a measure of the uncertainty of random variables. Given two continuous random variables $X$ and $Y$, the entropy of X is defined as:
\newline
\begin{equation}
    H(X) = - \int p(x)\:log\:p(x)\:dx
\end{equation}
The entropy of $X$ and $Y$ is defined as:
\newline
\begin{equation}
    H(X,Y) = - \iint p(x,y)\:log\:p(x,y)\:dx\:dy
\end{equation}
The MI between $X$ and $Y$ is defined as:
\newline
\begin{equation}
    MI(X;Y) = \iint p(x,y) \: log\:\frac{p(x,y)}{p(x)p(y)}\:dx\:dy
\end{equation}
where $p(x,y)$ is the joint probability density function of X and Y and $p(x)$ and $p(y)$ are the probability density function of X and Y respectively. 
\par
Hence MI and entropy can be combined in the form:
\begin{equation}
    MI(X;Y) = H(X)\:+\:H(Y)\:-H(X,Y)
\end{equation}
In this paper, the adjusted mutual information (AMI) score is used to calculate the correlations. The AMI score is an adjustment of the MI score to account for chance \cite{Vinh2010}. The AMI score is defined as:
\begin{equation}
    AMI(X,Y) = \frac{MI(X,Y) - E[MI(X,Y)]}{mean(H(X),H(Y))-E[MI(X,Y)]}
    \label{eqn:AMI}
\end{equation}
\subsection{Feature Selection for Time-Series Data}
A time series is a time based sequence of observations, $x_i(t); [i=1,\ldots, n; t=1,\ldots,m]$, where $i$ indexes the data gathered at time point $t$. The time series is univariate when $n$ is 1 and multivariate when $n$ is greater than or equal to 2. Multivariate time series can often be large in size and hence it is important to have suitable methods for preprocessing the data prior to classification. 
\par
To deal with the high dimensionality of MTS, two common methods used are feature extraction and feature subset selection. Feature extraction methods involve the transformation or mapping of the original data into extracted features. Feature subset selection involves reducing the number of features from the original dataset that is used for analysis by selecting only the features required and removing the redundant features.  One potential downfall of using feature extraction methods is that there can be a loss of information compared to using the original features. In this paper, the focus will be on feature subset selection methods.
\par
Many state of the art feature subset selection techniques such as Recursive Feature Elimination (RFE) require each item to be inputted in the form of a column vector \cite{IsabelleGuyon2003}. Multivariate time series tend to naturally be represented as a $m \times n$ matrix which makes these methods not ideal when working with multivariate time series for correlation based feature selection as vectorising time series data will lead to a loss of information about the correlation between the features. Hence, although there has been a lot of work undertaken in the area of multiple variable feature selection, there is limited work in feature selection for multivariate time series (MTS).
\par
Some correlation based methods have been implemented for feature subset selection in time series. Many methods typically used to calculate correlation such as Spearsman's correlation and rank correlation can be effective for non-time series data however has been shown to produce poor results when implemented on time series \cite{Wang2013}.
\par
Principal Component Analysis (PCA) is another technique that has been used in multivariate feature selection which allows correlation information between variables to be preserved. CLeVer is a technique which utilises properties of the descriptive common principal components for MTS feature subset selection. This method uses loadings to weight the contribution of each feature to the principal components. By ranking each feature by how much it contributes to the principal components, this method aims to reduce the dimensionality while retaining information related to both the original features and the correlation amongst the features \cite{Yanga}.
\par
Mutual Information (MI) is a popular technique that has been used on MTS data to measure correlation. MI is advantageous over other methods as it allows for both linear and nonlinear correlation to be captured. 
The class separability based feature selection (CSFS) algorithm uses MI between the original variables as features for classification. Based on this, the ratio of between class scattering to within class scattering is used to identify the contribution of a feature to the classification, hence allowing the original variables to be ranked according to their contribution to the classification \cite{Han2013}. 
\par
MI is generally calculated in a pairwise manner which may not be ideal when working with multidimensional data. To avoid this, some studies have used a k-nearest neighbour (k-NN) approach to calculate the MI which avoids the need to calculate the probability distribution function and therefore can be used on the original multidimensional feature subset \cite{Han2015,Liu2016}. Many of the methods using MI select the feature one by one using greedy search methods which may not lead to the identification of the optimal subset. The MSTS approach taken in this paper uses MI to evaluate correlation which is then used to calculate a merit score for each subset from which the best subset is selected.
\section{CFS for Time-Series Data}\label{sec:FS_TS}
CFS relies on the principle that ``\emph{a good feature subset is one that contains features highly correlated with the class, yet uncorrelated with each other}" \cite{hall1999correlation}. In this context where we use time series data, we aim to find a subset with features which are good predictors of the class while sharing little information with the other features in the subset.
\par
Typically, the correlation between the feature values themselves are calculated for use in CFS. As this is not feasible with time series data, we use the predictions of the class labels from each feature to help identify which features may be more correlated. The correlations could be defined in various ways including any distance measure between the feature-class and feature-feature class label predictions or through the use of mutual information based approaches. While investigating the best method to use to measure the correlations, initially the single feature accuracy was used for feature-class correlation and Hamming distance was used for feature-feature correlations. However, we decided to take a mutual information based approach for the correlations instead as it proved to give better accuracy. 
\par
\begin{figure*}
\centering
\includegraphics[trim = 70 30 70 30,clip, width = \textwidth]{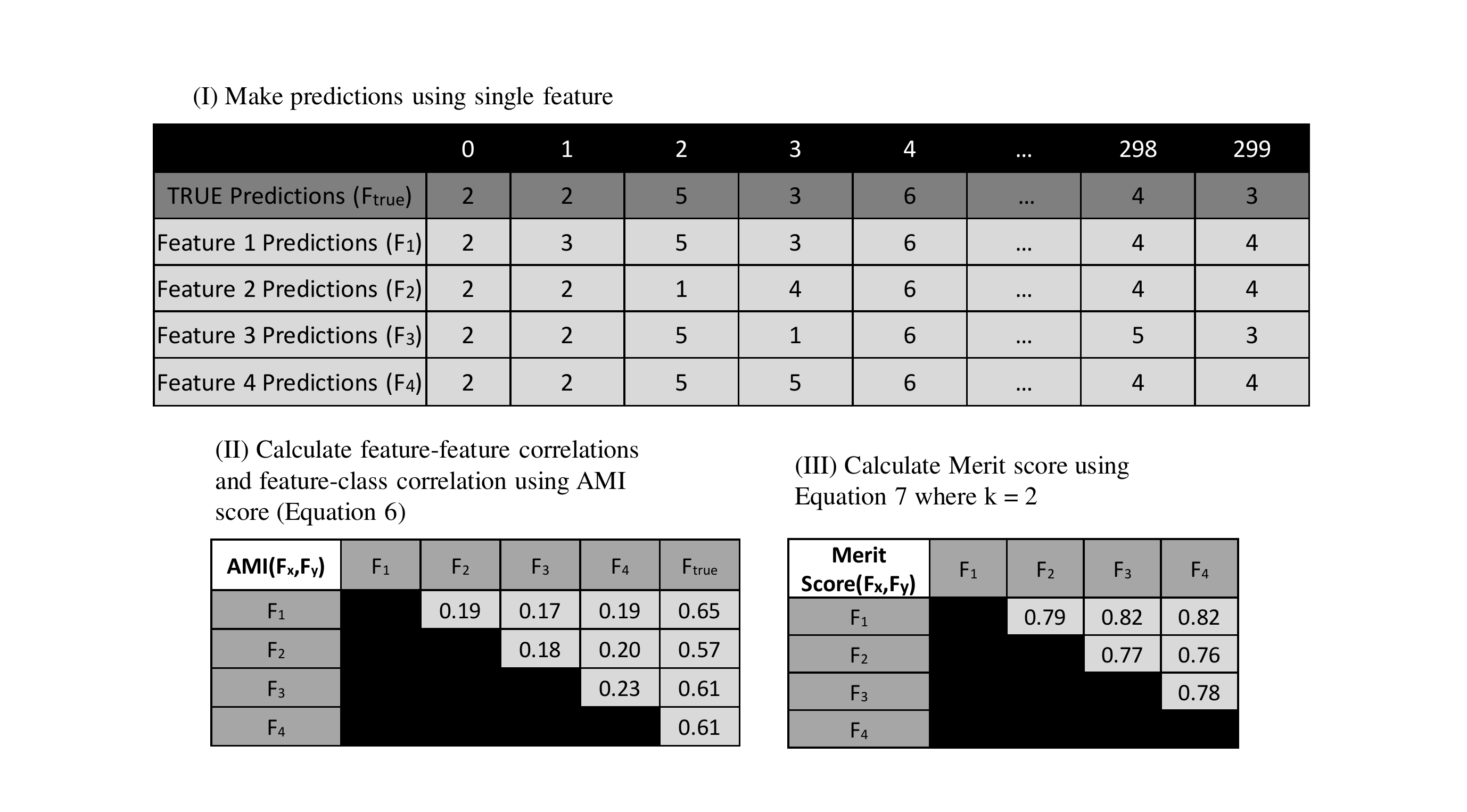}
\caption{Process taken to calculate the merit score of the ERing dataset for the 2-feature subset case. This is a six class problem with class labels 1 to 6.}\label{example_calc}
\end{figure*}
Figure \ref{example_calc} shows the process followed to calculate the MSTS where we initially make a prediction using each of the features separately. The predictions are then compared with the true labels using the Adjusted Mutual Information (AMI) score (Equation \ref{eqn:AMI}) to find the feature-class correlations and compared with the predictions of the other features, again using AMI to find the feature-feature correlations (Figure \ref{example_calc} (II)). Once the correlations have been identified, the merit score can be calculated using a modified version of Equation \ref{eqn:MS} as follows:
\begin{equation}
    MSTS = \frac{k\overline{Y_{cf}}}
    {\sqrt{k+k(k-1)\overline{Y_{ff}}}}
    \label{eqn:MSTS}
\end{equation}
where $Y_{cf}$ and $Y_{ff}$ are correlations calculated on the class labels predicted for the training data rather than on feature values as is the case in Equation \ref{eqn:MS}. Hence, $Y_{cf}$ was calculated by averaging the feature-class AMI score of all the features present in the subset. $Y_{ff}$ is calculated as the average of the pairwise AMI scores of each combination of features in the subset. The `best' subset would ideally be the one with the largest merit score. In this example there is a tie between $F_1,F_3$ and $F_1,F_4.$ - see Figure \ref{example_calc} (III).
\par
Following the merit score calculation, further evaluation is required to select the `best feature subset'. To do this, we evaluate two strategies. The strategies taken were:
\begin{enumerate}
    \item \textbf{Strategy 1}. The merit scores are calculated for all possible feature subset combinations (see section \ref{MS_eval}). The feature subset with the highest merit score was selected as the best feature subset.
    \item \textbf{Strategy 2}. Merit scores are calculated as in Strategy 1. The top 5 \% of the merit scores were selected and a Wrapper search was carried out on the selected feature subsets to identify the feature subset with the highest accuracy.
\end{enumerate}
Further detail on how this algorithm was evaluated is presented in the following section.
\section{Evaluation}\label{sec:eval}
In our evaluation we aim to assess the effectiveness of MSTS to identify good performing feature subsets and investigate how efficient this approach would be in terms of computational cost.
\subsection{Data Sets}
Six datasets were used for evaluation and these were all taken from the UEA multivariate time series classification archive \cite{Bagnall2018}. Five of these datasets are related to activity recognition and motion capture with one dataset from the audio spectra domain. All datasets were selected to have four or more dimensions. Four of the six datasets consist of accelerometer and/or gyroscope data. The ArticularyWordRecognition dataset has data obtained from an electromagnetic articulograph, a small sensor placed on the tongue and the JapaneseVowels dataset was taken from audio recordings. A summary of the datasets used for evaluation is shown in Table \ref{sumtable}.
\begin{table}
\centering
\begin{tabular}{ |x{5.2cm}|x{1.6cm}|x{1.6cm}|x{1.6cm}|x{1.6cm}| } 
\hline
 & Total \# of samples & \# of classes & \# of variables & Time series length \\ 
\hline
ArticularyWordRecognition (AWR) & 575 & 25 & 9 & 24\\ 
\hline
JapaneseVowels (JW) & 640  & 9 & 12 & 29\\
\hline
Cricket (Cr) & 180 & 12 & 6 & 1197\\
\hline
ERing (ER) & 60 & 6 & 4 & 65\\
\hline
NATOPS (NT) & 360 & 6 & 24 & 51\\
\hline
RacketSports (RS) & 303 & 4 & 6 & 30\\
\hline
\end{tabular}
\caption{Summary table of datasets used for evaluation}\label{sumtable}
\end{table}
\subsection{Merit Score Evaluation}\label{MS_eval}
\begin{figure}[t]
\centering
\includegraphics[scale=0.6]{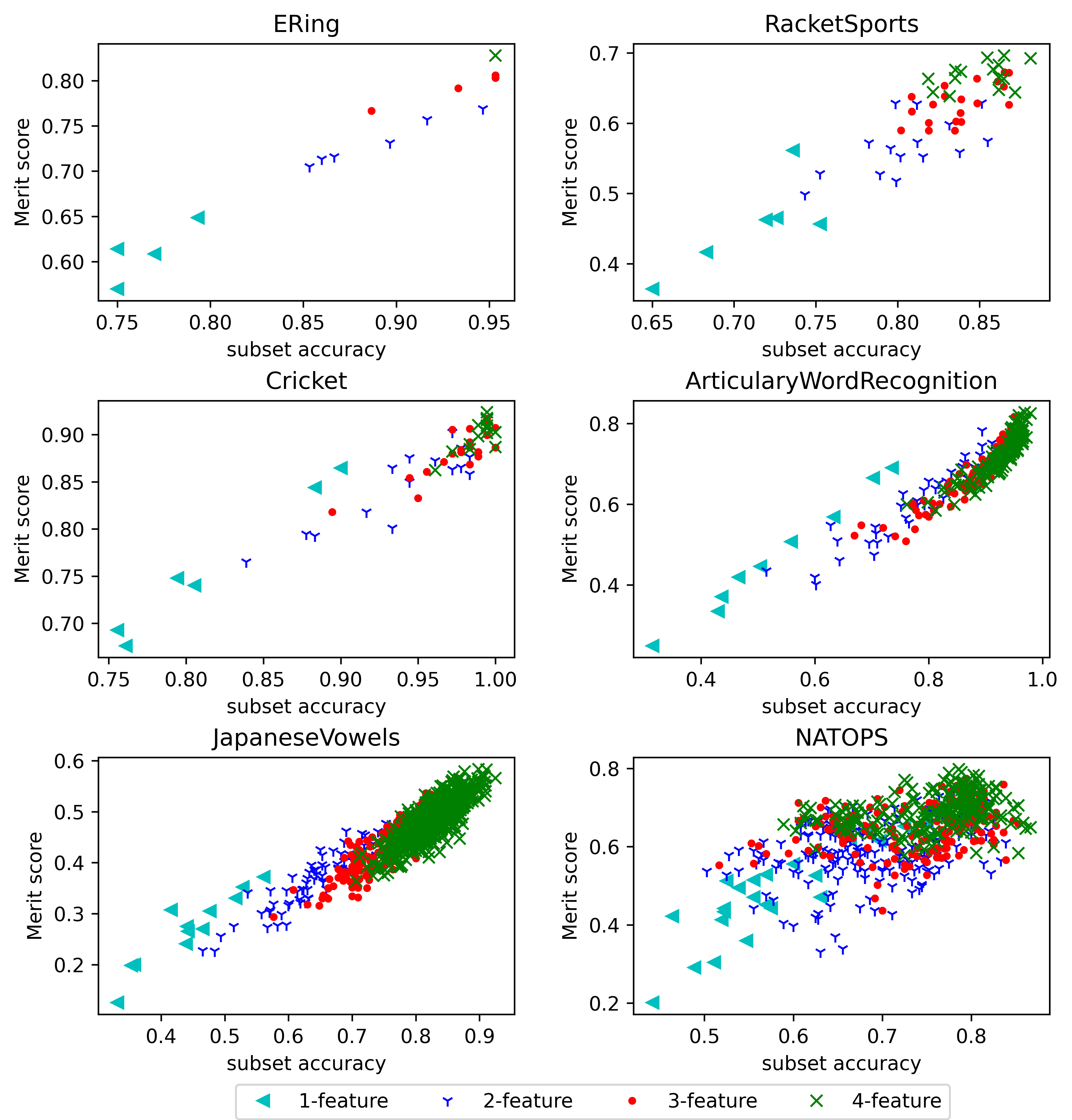}
\caption{MSTS vs subset accuracy for each feature subset combination for the six datasets}\label{acc_merit_all}
\end{figure}
The evaluation of the merit score was undertaken for feature subsets up to and including 4 features which was deemed sufficient as often MTS data only requires a small number of features to obtain high accuracy. To calculate the merit score for each dataset the following steps were taken:
\begin{enumerate}
    \item \textbf{Identify all unique feature subsets}. All unique combinations of feature subsets was identified and stored.
    \item \textbf{Calculate and store DTW distance matrix}. The similarity measure used for the time series in this paper is Dynamic Time Warping (DTW). DTW allows for a mapping of the time series in a non-linear way and works to find the optimal alignment between both series. DTW can be considered as a one-to-many mapping \cite{Sakoe1978}. As this is a computationally expensive task and will be repeatedly used for cross-validation, it is calculated and stored in advance.
    \item \textbf{Make class label predictions for each feature}. A 1-NN classifier using the stored DTW distances was used to do a 3-fold cross validation to make a set of class predictions using each feature individually.
    \item \textbf{Calculate feature-class and feature-feature correlations}. Calculate the feature-feature correlations and feature-class correlations as explained in Section \ref{sec:FS_TS}.
    \item \textbf{Calculate Merit Scores using Equation \ref{eqn:MSTS}.}
\end{enumerate}
\par
To compare the effectiveness of the merit score in identifying the optimal subsets, the classification accuracy of each subset was also calculated using a 1-NN-DTW classifier, which is often used as a benchmark technique whilst working with time series \cite{Bagnall2017}. A 3-fold cross validation was performed for each dataset. Figure \ref{acc_merit_all} shows the merit score against its subset accuracy for each feature subset.
\par
A positive trend is seen in Figure \ref{acc_merit_all} where a higher merit score generally corresponds to a higher accuracy. This trend is very visible in five out of six of the datasets with the NATOPS dataset yielding a less promising correlation in comparison with the other datasets. This behaviour may be due to the innate characteristics of the data which suggests that this approach may be more suitable for some datasets and domains than others. A slight feature subset size bias (SS-bias) is seen in the datasets where the different subsets sizes are forming clusters. However, overall the merit score gives a good indication of the better performing feature subsets and if the highest merit score was selected, a subset with a good classification accuracy would be selected as the `best' subset, although the optimal subset may not be selected. This is further evaluated in Section \ref{FSS}
\subsection{Feature Subset Selection}\label{FSS}
\begin{figure*}
\centering
\includegraphics[scale=0.43]{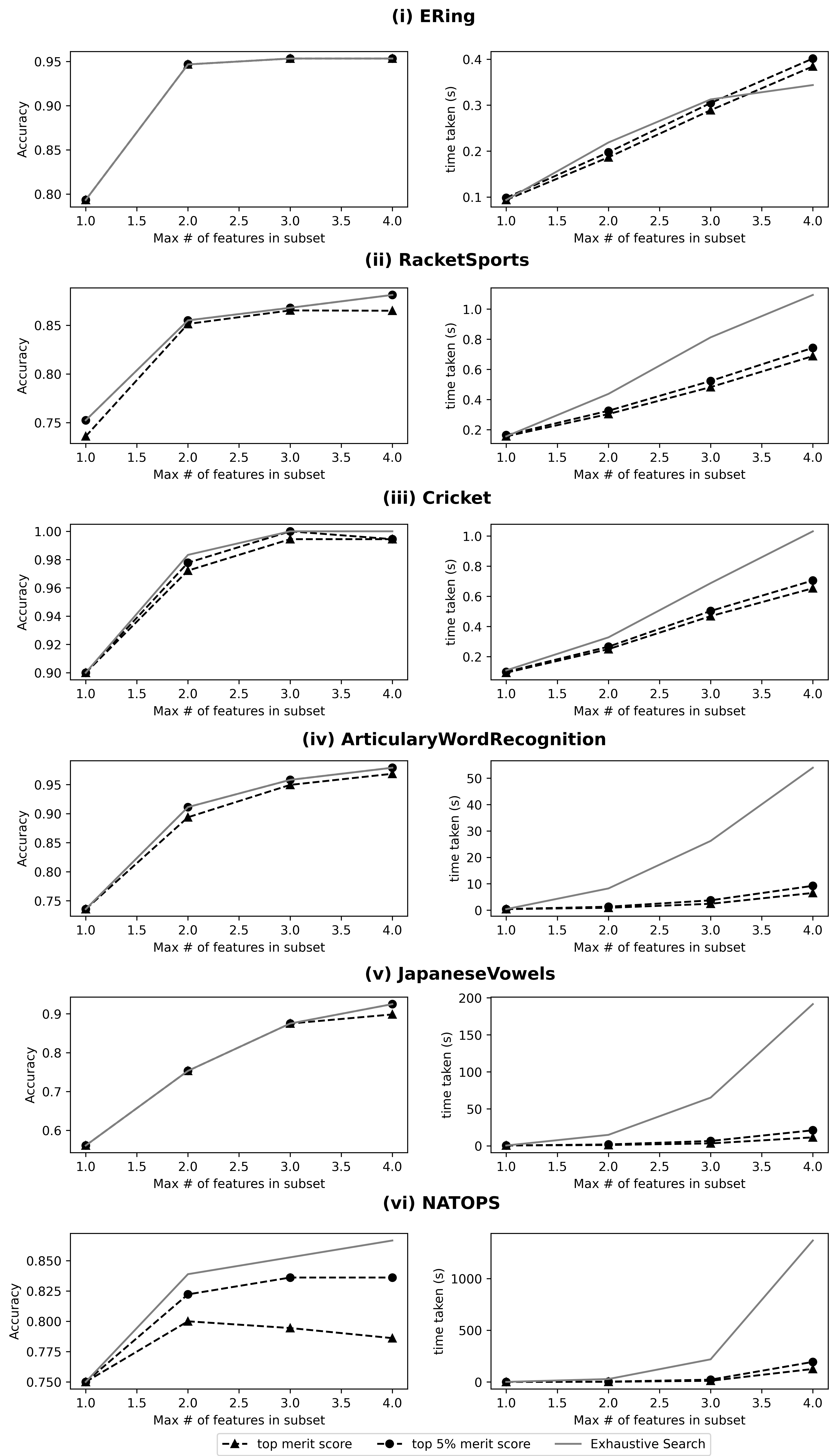}
\caption{Accuracy (Left) and computational time (Right) required for each of the datasets using the two MSTS strategies vs an exhaustive search}\label{acc_time_all}
\end{figure*}
Following the merit score calculation, the two strategies where we take the highest merit score to represent the best feature subset (Strategy 1) and we take the top 5\% of merit scores and undertake a search through this to find the best feature subset (Strategy 2) were both implemented on the datasets. For all evaluations of performance a 3-fold cross validation using 1-NN-DTW was used on the selected feature subsets. Figure \ref{acc_time_all} shows a comparison between the best accuracy and computational time required by the two strategies undertaken using the merit score and compares this with that from an exhaustive search through all unique feature combinations. The computational time recorded for the two MSTS approaches includes the calculation of the merit score itself and the 1-NN-DTW search using either the best feature subset or through all feature subsets which belong to the top 5\% of the merit scores. The computational time recorded for the exhaustive search includes the calculation of accuracy for all unique feature subsets. The unique feature subsets possible for each dataset and the DTW distance matrices are both calculated and saved in advance as they are common to both approaches, hence they have not been included in the computational time calculations.
\par
From the results it can be seen that for most cases, the best subset was able to be obtained using the MSTS strategy 2 where the top 5\% of merit scores were used. Although strategy 1 is also able to obtain a good accuracy in most cases, the best feature subset is only found using this strategy for the ERing dataset. The exception where the MSTS strategies did not work perfectly was in the NATOPS dataset where the best accuracy obtained was about 4-5\% less using the MSTS and the Cricket dataset where the best accuracy obtained was less than 1\% below the optimal accuracy. The reasoning for the undesirable performance of NATOPS can be seen in Figure \ref{acc_merit_all} where the NATOPS 4 variable subset has a less positive relationship between merit score and subset accuracy in comparison with the other datasets. The computational time required for the identification of the best subset was faster using MSTS for all except the ERing dataset. As ERing had the smallest number of dimensions (4 dimensions) this is not a surprising result. As the number of dimensions increase it is evident that MSTS is highly suitable to reduce computational cost as the larger datasets such as NATOPS (24 dimensions), JapaneseVowels (12 dimensions), and articularyWordRecognition (9 dimensions) see a large reduction in time taken while using this approach. As the time difference between the two MSTS strategies are minimal, strategy 2 where the top 5\% of all merit scores are evaluated performs best overall giving near perfect performance identification in 5 out of the 6 datasets.
\section{Conclusions \& Future Work}
In this paper, a feature subset selection technique based on merit scores is implemented for multivariate time series. The technique employed here uses correlations based on classifiers from single features to identify a subset with low feature to feature correlation and high feature-class correlations. The evaluation carried out in this paper suggests that this approach can lead to a considerable reduction in the computational time required to identify a good subset. This approach is in particular useful for very high dimension data as the reduction in computational time by MSTS improves as the number of dimensions increases. 
\par
The results suggest good potential for this approach to be used as a feature selection technique for time series as a high accuracy yielding subset was selected in each of the datasets that were evaluated. Of the datasets analysed, near optimal results were obtained for five of the six datasets. Hence, the question of whether the nature of the data impacts the effectiveness of the approach is still unanswered. This will be investigated in the future with the aim of getting a better understanding of under what conditions this technique will be most effective. 

To deal with very high dimension datasets, in our future work we will attempt a greedy search through the features for the merit score calculation rather than calculating the merit score for all subset combinations. Another direction for further work is to explore the effectiveness of correlations based on subsets of the available data, e.g. 100 samples with the aim of reducing the amount of training data required to carry out the feature selection.
\section*{Acknowledgements}
This work was funded by Science Foundation Ireland through the SFI Centre for Research Training in Machine Learning (Grant No. 18/CRT/6183)
\bibliographystyle{abbrv}  
\bibliography{aaltd}
\end{document}